\newcommand{\drift}{\textsc{DRIFT}}
\newcommand{\heteroprompt}{\textsc{HeteroPROMPT}}
\documentclass[letterpaper, 10 pt, conference]{ieeeconf} 

\IEEEoverridecommandlockouts    
\makeatletter
\def\ps@copyrightnotice{%
  \def\@oddhead{}%
  \def\@evenhead{}%
  \def\@oddfoot{%
    \hfil
    \parbox[t]{0.92\textwidth}{%
      \centering
      \footnotesize
      \textcopyright{} 2026 IEEE. Personal use of this material is
      permitted. Permission from IEEE must be obtained for all other
      uses, in any current or future media, including
      reprinting/republishing this material for advertising or
      promotional purposes, creating new collective works, for resale
      or redistribution to servers or lists, or reuse of any copyrighted
      component of this work in other works.%
    }%
    \hfil
  }%
  \let\@evenfoot\@oddfoot
}
\makeatother

\usepackage{graphics} 
\usepackage{epsfig} 
\usepackage{mathptmx} 
\usepackage{times} 
\usepackage{amsmath} 
\usepackage{amssymb}  
\usepackage{cite}
\usepackage{color, soul}
\usepackage{amsfonts}
\usepackage{wasysym}
\usepackage{booktabs}
\usepackage{algorithm}
\usepackage{algpseudocode}

\algnewcommand{\Input}[1]{\Statex \textbf{Input:} #1}
\algnewcommand{\Output}[1]{\Statex \textbf{Output:} #1}

\algrenewcommand{\algorithmiccomment}[1]{%
    \hfill\texttt{// #1}%
}\usepackage{url}
\usepackage{pifont}

\newcommand{\Crossmark}{\ding{55}}
\title{\LARGE \bf
\heteroprompt: A Real-time and Privacy-Preserving Heterogeneous Collaborative Perception Framework
}
\author{Armin Maleki$^{1}$ (malekiar@msu.edu) and Hayder Radha$^{1}$ (radha@msu.edu) \\ $^{1}$ Electrical and Computer Engineering, Michigan State University}
\begin{document}
\maketitle
\thispagestyle{copyrightnotice}

\pagestyle{empty}

\begin{abstract}
Collaborative Perception (CP) improves autonomous systems' awareness of their surroundings via shared data, intermediate features, and/or detection results.
However, in real-world scenarios, collaborating vehicles often use heterogeneous sensors, perception models, datasets, and training domains, causing a domain shift in the feature space, which degrades downstream fusion and detection. Previous efforts aligned this domain gap by retraining fusion and detection components or by introducing feature interpreters for each modality, which scale poorly to new joining agents and mostly rely on access to proprietary metadata, raising privacy concerns.
In this paper, we propose \textbf{\heteroprompt},
a heterogeneous, real-time, and privacy-preserving collaborative perception framework. \textbf{\heteroprompt}~supports rapid heterogeneous agents' features alignment using modular prompts and learning-based tuning, which aligns each heterogeneous agent modality to an ego-centric unified feature space, while freezing agents' encoders, and collaboration fusion and detection stacks.
Our novel framework performs visual prompt-based training and inference to efficiently modulate Bird's Eye View (BEV) features across channels and spatial locations while significantly reducing computational overhead. To enable a metadata-free deployment, we use an AutoEncoder to learn a compact unified feature space, extracting modality cues from shared features to support real-time and privacy-preserving modality classification and routing to the appropriate \heteroprompt's modules.
Experiments on the OPV2V-H and V2XSet datasets show that our framework improves object detection performance, in terms of Average Precision (AP), over state-of-the-art heterogeneous CP methods while using orders of magnitude fewer trainable parameters, representing a scalable, practical, and privacy-preserving solution to heterogeneous CP.
Moreover, the proposed novel modality classification scheme rapidly predicts the joining agent's modality from compact features with $>99.99\%$ accuracy during deployment. \textnormal{Code is available at~\url{https://github.com/arminmaleki007/HeteroPROMPT}.}
\end{abstract}

\section{INTRODUCTION}
Collaborative Perception (CP) improves connected vehicles' awareness of their surroundings by enabling agents to share perception data, thereby supporting safer driving decisions than single-agent perception, which is limited by a restricted field of view and occlusions~\cite{b1}. Despite rapid progress, CP systems still suffer from key bottlenecks, including bandwidth constraints~\cite{b2,b3}, sensitivity to communication latency~\cite{b4,b5}, pose noise/uncertainty~\cite{b6,b7}, and feature-domain shifts arising from heterogeneous sensors and perception models. 
The majority of existing approaches address subsets of these issues but assume homogeneous sensors and models across collaborating agents. Such an assumption imposes strong deployment constraints: performance can degrade sharply when a newly observed neighbor uses a different sensor, model, or training data than those seen during training. In practice, vehicles employ diverse and often proprietary perception stacks that introduce a domain gap among intermediate features, which leads to degradation in both perception accuracy and driving safety. Therefore, aligning this domain gap to enable robust collaboration is essential for deploying CP in real-world autonomous driving.

A single fusion model with fixed weights has not yet been developed to bridge the feature-domain gap induced by diverse sensor and model configurations encountered in practice. As a result, prior work typically adapts the collaboration pipeline to a limited set of heterogeneous neighbors, either by (i) retraining major components of the fusion and detection models~\cite{b8,b9,b10,b11} or the joining agent's backbone~\cite{b12}, or by (ii) attaching an extra feature interpreter that aligns intermediate features for each new agent~\cite{b13,b14,b15}.
Approaches in the first category often incur substantial retraining costs when new modalities appear and may raise privacy concerns because they require updating the joining agent's model, and hence revealing details of the model. For example, HEAL~\cite{b12} constructs a homogeneous unified base and then adapts newly joining heterogeneous encoders, leading to repeated retraining and requiring access to the joining agent's model parameters. In contrast, feature-interpreter-based approaches can better preserve privacy but may rely on more complex models to align the features effectively, which can slow down deployment.

Nevertheless, many existing approaches rely on proprietary sensor and model metadata to select the appropriate adaptation module (model/interpreter) for each joining agent during deployment. In practice, sharing such metadata is often infeasible because perception stacks are companies' proprietary and disclosure may expose sensitive design details or increase the attack surface. These limitations motivate a lightweight and privacy-preserving framework that can select the appropriate adaptation modules for each joining agent \emph{in real time} without requiring any proprietary metadata, while maintaining detection accuracy and computational efficiency.

In this work, we propose \textbf{\heteroprompt}, a \textbf{Hetero}geneous collaborative perception framework for real-time, \textbf{P}rivacy-preserving, and \textbf{R}apid \textbf{O}bject-alignment using \textbf{M}odular \textbf{P}rompts and \textbf{T}uning that addresses BEV feature-domain gaps among heterogeneous agents. In \textbf{Stage~1}, we follow prior efforts (e.g., HEAL~\cite{b12}) to learn an ego-centric unified feature space under a homogeneous setting. To support heterogeneous deployment, we introduce \textsc{DRIFT} (\textbf{D}ecomposed-\textbf{R}ank \textbf{I}nterpreter for \textbf{F}iLM \textbf{T}uning), a modality-specific interpreter that aligns a newly joining agent's intermediate features to the unified space. In \textbf{Stage~2}, we fine-tune \textsc{DRIFT} at two points in the detection pipeline to efficiently reduce the domain gap for each new modality while keeping the ego fusion/head and the joining agent encoder frozen. We further train an AutoEncoder (within an intermediate Stage~1.5, between stages 1 and 2) to obtain a compact unified feature space using the frozen homogeneous base, which will be used to infer the joining agent's modality in real-time deployment without requiring any proprietary sensor/model metadata. \heteroprompt~preserves the single-agent perception stack (for when no collaboration exists) and the privacy of newly joining agents during both training and deployment, while enabling scalable adaptation with low computational and storage overhead.
 
\textsc{DRIFT} implements Feature-wise Linear Modulation (FiLM)~\cite{b18} using low-rank visual prompts to adaptively modulate intermediate BEV features across channels and spatial locations. To reduce the number of trainable parameters and improve optimization efficiency, we apply a Parallel Factor (PARAFAC~\cite{b26}) low-rank decomposition to the prompt tensors, substantially lowering the adaptation complexity compared to dense prompts and accelerating convergence while retaining competitive detection performance.

Our key contributions are summarized as follows:
    \begin{itemize}
    \item We propose \textbf{\heteroprompt}, a lightweight heterogeneous collaborative perception framework that aligns newly joining agents to an ego-centric unified feature space. For each new modality, we adapt only two learnable visual prompt-conditioned FiLM modules, requiring access solely to communicated intermediate features while keeping sensor/model configurations private.
    \item We introduce a compact representation of the ego unified feature space by training an AutoEncoder, and learn modality cues from BEV intermediate features in this compact space. Then, we train a classifier using these modality cues. At deployment, the ego agent uses this feature-based classifier to predict the joining agent’s modality and select the appropriate prompt--FiLM pair without relying on any proprietary metadata.
    \item We reduce adaptation cost by using PARAFAC low-rank decomposition for visual prompts, enabling fast feature alignment with substantially fewer trainable parameters, lower computation cost, and reduced storage by saving only modality-specific \textsc{DRIFT} modules.
    \item Extensive experiments demonstrate that \heteroprompt~achieves state-of-the-art detection accuracy, while reducing average training overhead by $96\%$, and hence supporting scalable and privacy-preserving deployment of collaborative perception in real-world autonomous driving. In addition, the proposed modality classification scheme classifies the compact features during deployment with $>99.99\%$ accuracy.

    \end{itemize}
\section{RELATED WORKS}

\subsection{Collaborative Perception}
Collaborative perception (CP) enables autonomous vehicles to improve environmental awareness by sharing perceptual information, particularly under occlusions and limited field-of-view~\cite{b1}. Early CP approaches perform \emph{early fusion} by transmitting raw sensor data, which can achieve strong performance but incurs substantial bandwidth cost. To reduce communication load, \emph{late fusion} shares only high-level detection outputs, offering low bandwidth cost but often being limited by local prediction errors. More recent methods adopt \emph{intermediate fusion}, where agents exchange intermediate features to balance bandwidth and detection accuracy. 
Beyond fusion strategies, prior work has investigated communication-efficient feature compression and bandwidth allocation~\cite{b2,b3}, robustness to latency, and communication~\cite{b4,b5} and detection 
noise/uncertainty~\cite{b6,b7}. However, performance degradation caused by feature-domain shifts among heterogeneous agents (e.g., different sensors, models, or training domains) remains comparatively underexplored, despite being critical for safe and reliable real-world CP.

\subsection{Heterogeneous Collaborative Perception}
Aligning heterogeneous intermediate features is critical for collaborative perception when agents employ different sensors, models, datasets, or training domains; otherwise, feature misalignment can significantly degrade fusion performance. Existing approaches generally fall into two categories. The first category reduces the domain gap by retraining core components, such as the fusion/detection modules~\cite{b8,b9,b10,b11} and/or the joining agent's encoder. For example, HEAL~\cite{b12} constructs an ego-centric unified space and then adapts heterogeneous encoders to match it. Although effective, such retraining-based solutions are computationally expensive for newly emerging modalities and raise privacy concerns because they require access to the joining agent's model parameters for optimization.
The second category introduces model interpreters that align heterogeneous features while freezing the main collaborative pipeline and agent encoders~\cite{b13,b14,b15}. These methods can better preserve privacy and enable faster adaptation, but often introduce additional complexity and may incur non-trivial inference latency. Recent prompt-based variants reduce this cost using lightweight prompts. For example, Faster-HEAL~\cite{b16} uses low-rank prompts with a learnable aligner, while PEARL~\cite{b17} performs anonymous real-time selection using modality-specific compressors/interpreters and domain-invariant feature similarity.
In contrast, \heteroprompt~performs metadata-free routing through a compact AutoEncoder (AE)-based latent space and a lightweight classifier, using a shared resizer before modality inference. This routes communicated BEV features to the corresponding prompt-conditioned FiLM/DRIFT modules without requiring proprietary metadata or modality-specific learnable compression for the routing stage.

\subsection{Feature Modulation}
A practical strategy for domain adaptation is to modulate intermediate activations with lightweight, conditional transformations, instead of retraining large perception backbones. Conditional normalization methods achieve this by predicting per-channel affine parameters conditioned on an external signal, e.g., (CBN)~\cite{b19}. Related channel-wise recalibration mechanisms, such as squeeze-and-excitation (SE) blocks~\cite{b20}, gate feature channels to improve robustness and representation capacity with minimal additional parameters. These approaches demonstrate that channel-wise modulation can efficiently reshape feature distributions while keeping the main network largely unchanged.
Building on this family, Feature-wise Linear Modulation (FiLM)~\cite{b18} provides a general conditioning layer that applies learnable, feature-wise affine transformations to intermediate representations, making it a natural fit for parameter-efficient feature alignment. In parallel, prompt-based learning adapts models by optimizing small learnable visual prompts while freezing the backbone, as in 
low-rank prompting~\cite{b21}. Prompting has also been explored for detection and domain shift, including visual prompt-driven domain generalization/adaptation and detection-oriented settings~\cite{b22,b23,b24}. Motivated by these advances, we combine visual prompts with FiLM-style channel-wise modulation to obtain a compact interpreter for aligning heterogeneous intermediate BEV features, while keeping the main collaborative perception pipeline frozen.

\section{METHODOLOGY}
\begin{figure*}
    \centering
\includegraphics[width=0.94\linewidth]{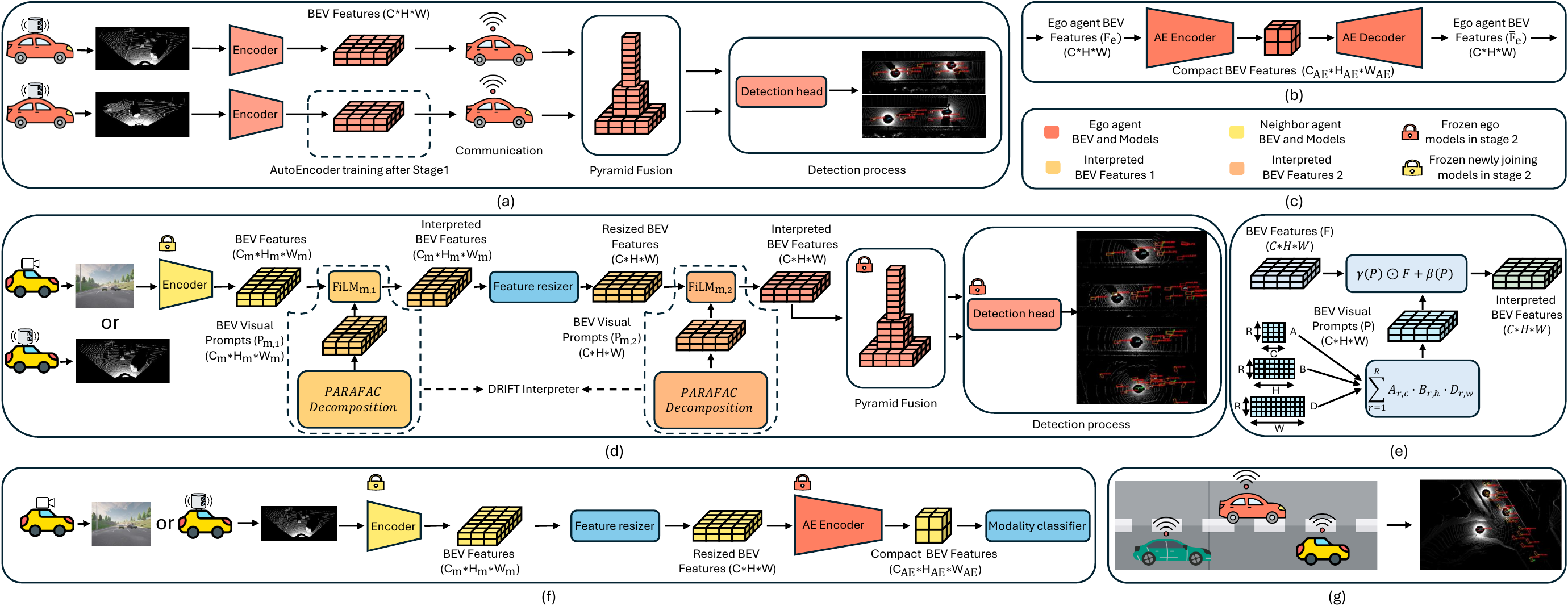}
      \caption{\textbf{Overview of \heteroprompt}. (a) Stage~1 learns an ego-centric unified BEV feature space for collaborative fusion and detection. (b) Stage~1.5 trains an AE to obtain a compact latent representation from ego-domain BEV features. (c) Color/marker legend. (d) Stage~2 fine-tunes two \textsc{DRIFT} modules to align a newly joining heterogeneous agent to the unified space. (e) \textsc{DRIFT} uses PARAFAC low-rank prompt-conditioned FiLM to modulate BEV features. (f) Metadata-free modality inference maps communicated BEV features to the AE latent space and classifies the modality for \textsc{DRIFT} selection. (g) Heterogeneous CP after routing and feature alignment.}
    \label{fig:overall}
\end{figure*}

Fig.~\ref{fig:overall} illustrates the overall architecture of \heteroprompt. We adopt a two-stage training paradigm with an intermediate AutoEncoder (AE) training step (Stage~1.5), to address feature misalignment in heterogeneous collaborative perception. \textbf{Stage~1} learns an ego-centric unified feature space for detection by following the homogeneous base architecture and loss design of HEAL~\cite{b12} (summarized in Subsection~\ref{subsec:homogeneous-base}). To obtain a compact representation for deployment, we further introduce a shallow distillation step via an AutoEncoder, which is used at inference time. \textbf{Stage~2} introduces and fine-tunes two lightweight visual prompts beside feature-wise linear modulation to bridge the domain gap between heterogeneous intermediate features and the ego agent unified space, enabling fast adaptation with low inference overhead while preserving privacy and single-agent performance~\footnote{Here, \textit{single-agent} performance refers to the performance when an agent has to revert to its own perception results without the benefits of other collaborating agents' features e.g., due to interruption in connectivity.},
unlike retraining. Two prompts operate at different points in the detection pipeline, aligning BEV features to the unified space in a complementary manner.

\subsection{Stage~1: Homogeneous Unified Space Training}
\label{subsec:homogeneous-base}
The purpose of Stage~1 is to learn an ego-centric unified feature space under a homogeneous setting, where all agents share the same encoder. Let $\mathcal{A}=\{1,\dots,N\}$ denote the agent set including the ego agent $e$ and neighbors $j\in\{\mathcal{A} \backslash \{e\}\}$. Each agent $i\in\mathcal{A}$ encodes its observation $O_i$ into a BEV feature map $F_i$ using the shared encoder $f_{m_1(ego)}$, which would be transmitted to the ego agent through a communication operator $\Gamma_{i\rightarrow e}$. Here, \textit{modality} represents the collective configuration of an agent's sensors, encoder, and detection architecture/model. Hence, $m_1$ represents the ego agent modality, and $f_{m_1(ego)}$ is the ego agent unified feature space encoder function. The ego agent aggregates the received features via pyramid fusion to obtain the unified feature representation $H_{m_1}$, which is then fed into the detection head to predict final detections $B$, described in Algorithm~\ref{alg:stage1_forward}.
\begin{algorithm}[t]
\caption{Stage~1 Training Logic for Agent Set $\mathcal{A}$}
\label{alg:stage1_forward}
\begin{algorithmic}[1]

\Input{Agent observations $\{O_i\}_{i\in\mathcal{A}}$.}
\Output{Detection output $B$.}

\Statex \textbf{1.~Per-agent feature extraction:}
\State $F_i \gets f_{m_1(\text{ego})}(O_i),\quad i\in\mathcal{A}$
\Comment{Shared encoder}

\Statex \textbf{2.~Feature communication:}
\State $F_{i\rightarrow e}
\gets \Gamma_{i\rightarrow e}(F_i),\quad i\in\mathcal{A}$
\Comment{Communication}

\Statex \textbf{3.~Fusion and detection:}
\State $H_{m_1}
\gets f_{\text{pyramid}}
\left(F_{1\rightarrow e},\dots,F_{N\rightarrow e}\right)$
\Comment{Pyramid fusion}

\State $B \gets f_{\text{head}}(H_{m_1})$
\Comment{Detection head}

\State \Return $B$

\end{algorithmic}
\end{algorithm}

\heteroprompt~adopts a pyramid fusion scheme for efficient and accurate fusion, following~\cite{b12}. We perform fine-to-coarse fusion by downsampling the received features using a ResNeXt~\cite{b25} block to obtain $L$ scales. At each scale $l$, a foreground estimator $f_{\text{fg}}^{(l)}$ produces an occupancy map to highlight informative regions. Fusion weights are computed by applying a softmax across agents. The fused features are then upsampled to a common resolution and concatenated to produce $H_{m_1}$, described in Algorithm~\ref{alg:pyramid_fusion}.
\begin{algorithm}[t]
\caption{Pyramid Fusion Logic}
\label{alg:pyramid_fusion}
\begin{algorithmic}[1]

\Input{Aligned features
$\{F_{i\rightarrow e}^{(0)}\}_{i=1}^{N}$.}
\Output{Fused pyramid feature $H_{m_1}$.}

\Statex \textbf{1.~Multi-level feature extraction and weighting:}

\State $F_{i\rightarrow e}^{(l)}
\gets f_{\text{ResNeXt}}^{(l)}
\left(F_{i\rightarrow e}^{(l-1)}\right),
\quad l\in L,\ i\in\mathcal{A}$

\State $OCC_i^{(l)}
\gets f_{\text{fg}}^{(l)}
\left(F_{i\rightarrow e}^{(l)}\right)$
\Comment{Foreground score}

\State $W_{1:N}^{(l)}
\gets \operatorname{softmax}
\left(OCC_1^{(l)},\dots,OCC_N^{(l)}\right)$
\Comment{Normalize}

\State $F_{m_1}^{(l)}
\gets \displaystyle\sum_{i=1}^{N}
F_{i\rightarrow e}^{(l)}\cdot W_i^{(l)}$
\Comment{Weighted sum}

\State $F_{m_1}^{(l)}
\gets f_{\text{upsample}}^{(l)}
\left(F_{m_1}^{(l)}\right)$
\Comment{Match resolution}

\Statex \textbf{2.~Multi-scale concatenation:}

\State $H_{m_1}
\gets \operatorname{concat}
\left(
F_{m_1}^{(1)},
F_{m_1}^{(2)},
\dots,
F_{m_1}^{(L)}
\right)$

\State \Return $H_{m_1}$

\end{algorithmic}
\end{algorithm}

Besides the final detection loss $\mathcal{L}_{\text{det}}$, which includes focal classification loss $\mathcal{L}_{\text{focal}}$, regression loss $\mathcal{L}_{\text{smooth-L1}}$, and direction loss $\mathcal{L}_{\text{dir}}$, Stage~1 additionally supervises the foreground estimator at each pyramid level using a focal loss:
\begin{equation}
\label{eq:loss}
    \begin{gathered}
        \mathcal{L} = \mathcal{L}_{\text{det}}(B,Y) + \sum_{l=1}^{L} \alpha_l \, \mathcal{L}_{\text{foreground}}^{(l)} \\
        \mathcal{L}_{\text{det}}(B,Y) = \mathcal{L}_{\text{focal}}(B,Y)\\ 
        + 2 \cdot \mathcal{L}_{\text{smooth-L1}}(B,Y)
        + 0.2 \cdot \mathcal{L}_{\text{dir}}(B,Y) \\
        \mathcal{L}_{\text{foreground}}^{(l)} = \sum_{i=1}^{N} \mathcal{L}_{\text{focal}}\!\left(OCC_{i}^{(l)},Y_{i}^{(l)}\right),
    \end{gathered}
\end{equation}
where $Y$ is the ground-truth, $B$ denotes the  predictions, $OCC_{i}^{(l)}$ is the predicted occupancy map, and $Y_{i}^{(l)}$ is the foreground mask with $\alpha_l$ as weighting contribution of scales.
\subsection{Stage~2: New Joining Heterogeneous Agent}
After Stage~1 learns an ego-centric unified feature space, new agents may join the collaboration group. Since the unified space is learned under the ego agent's sensor/model configuration, a distribution mismatch in BEV features arises when a new agent employs different sensors, perception backbones, or datasets, which can significantly degrade the detection performance. Therefore, the new agent's features must be aligned to the unified feature space to minimize performance loss. 
In practice, retraining the fusion and detection modules for every newly joining agent (or each new agent backbone for each base) is often infeasible. This motivates a lightweight and privacy-preserving adaptation mechanism that can efficiently align heterogeneous features under constraints such as limited computation/resources and the proprietary nature of companies' sensors and models.

\heteroprompt~addresses these constraints by freezing the ego fusion module and detection head, thereby preserving the Stage~1 unified feature space. We also freeze the newly joining agent's encoder, confining adaptation to a small set of prompts operating on shared intermediate features; this reduces computational overhead and better preserves the new agent's privacy. By significantly reducing complexity and by preserving critical aspects of new agent's metadata (i.e., sensor type, encoder, detection head model, etc.), our approach represents a major departure from prior CP frameworks, including HEAL~\cite{b12}. 
To align heterogeneous features to the unified space, we introduce \textbf{D}ecomposed-\textbf{R}ank \textbf{I}nterpreter for \textbf{F}iLM \textbf{T}uning (\textbf{DRIFT}) at two points in the detection pipeline containing lightweight and modality-specific visual prompts, and fine-tune them. For a newly joining agent of modality $m_2$, we use a set of agents $\mathcal{A}_{m_2}$ and a frozen encoder $f^{**}_{enc_{m_2}}$. The alignment procedure is described in Algorithm~\ref{alg:stage2_forward}:

\begin{algorithm}[t]
\caption{Stage 2 Training Logic for Joining Agent $m_2$}
\label{alg:stage2_forward}
\begin{algorithmic}

\Input{Joining agent observation
$\{O_{k,m_2}\}_{k\in\mathcal{A}_{m_2}}$}

\Output{Detection output $B_{m_2}$}

\Statex \textbf{1. Frozen feature extraction and communication:}

\State $F_{k,m_2}
\gets f_{\mathrm{enc}_{m_2}}^{**}(O_{k,m_2}),
\ k\in\mathcal{A}_{m_2}$
\Comment{Frozen encoder}

\State $F_{k,m_2\rightarrow e}
\gets \Gamma_{k,m_2\rightarrow e}(F_{k,m_2}),
\ k\in\mathcal{A}_{m_2}$ \Comment{Communication}

\Statex \textbf{2. \drift{} fine-tuning:}

\State $F'_{k,m_2}
\gets FiLM_{m_2,1}
(F_{k,m_2\rightarrow e},P_{m_2,1})$
\Comment{\drift{} 1 tuning}

\State $F''_{k,m_2}
\gets f_{\mathrm{resizer}}(F'_{k,m_2})$
\Comment{Match ego-agent size}

\State $F'''_{k,m_2}
\gets FiLM_{m_2,2}
(F''_{k,m_2},P_{m_2,2})$
\Comment{\drift{} 2 tuning}

\Statex \textbf{3. Frozen fusion and detection:}

\State $H_{m_1,m_2}
\gets f_{\mathrm{pyramid}}^{*}
\left(\{F'''_{k,m_2}\}_{k\in\mathcal{A}_{m_2}}\right)$
\Comment{Fusion}

\State $B_{m_2}
\gets f_{\mathrm{head}}^{*}(H_{m_1,m_2})$
\Comment{Detection head}

\State \Return $B_{m_2}$

\end{algorithmic}
\end{algorithm}

Here, $F_{k,m_2}$ is the BEV feature produced by the new agent k, $F_{k,m_2 \rightarrow e}$ denotes the communicated BEV feature after synchronization and ego-frame transformation at the current collaboration timestamp, following the benchmark CP pipeline, and $P_{m_2,i}$ is the learnable prompt for modality $m_2$ at step $i$. $f_{\text{resizer}}$ maps the new agent feature to the common unified resolution (e.g., spatial resizing and lightweight channel projection). The superscripts $*$ (Stage~1) and $**$ (new agent) indicate frozen pretrained modules during Stage~2. We use \textbf{F}eature-w\textbf{i}se \textbf{L}inear \textbf{M}odulation (FiLM)~\cite{b18}, defined as:
\begin{equation}
    FiLM_{m,i}(F_{m},P_{m,i}) = \gamma_{m,i} (P_{m,i}) \odot F_{m} + \beta_{m,i}(P_{m,i}), 
\end{equation}
where $\gamma_{m,i}(\cdot)$ and $\beta_{m,i}(\cdot)$ are produced from the prompt $P_{m,i}$ for the modality $m$ at step $i$ using a lightweight MLP or CNN, and are broadcasted to match the feature dimensions.

To ensure compatibility, the first prompt $P_{m_2,1}$ matches the shape of the communicated feature, i.e., $(C_{m_2},H_{m_2},W_{m_2})$, while the second prompt $P_{m_2,2}$ matches the unified resolution after resizing, i.e., $(C,H,W)$. Directly optimizing dense prompts of size $C\times H\times W$ can introduce a large number of trainable parameters, increasing optimization difficulty and slowing convergence. 
To reduce the trainable parameter count, we apply PARAFAC (CP) decomposition~\cite{b26} to represent each prompt with a low-rank factorization of rank $R$. Specifically, we learn three factor matrices $A\in\mathbb{R}^{R\times C}$, $B\in\mathbb{R}^{R\times H}$, and $D\in\mathbb{R}^{R\times W}$, and reconstruct the prompt as
\begin{equation}
\label{equ:cp}
    P_{c,h,w} \approx \sum_{r=1}^{R} A_{r,c} \cdot B_{r,h} \cdot D_{r,w},
\end{equation}
where $P_{c,h,w}$ denotes the prompt element at channel $c$ and spatial location $(h,w)$. This decomposition reduces the number of trainable parameters from $C\times H\times W$ to $R(C+H+W)$ with $R\ll \min(C,H,W)$, lowering training cost and accelerating convergence while maintaining comparable performance. 

FiLM~\cite{b18} is used to modulate intermediate BEV features using the learnable prompts, thereby aligning heterogeneous features to the ego agent unified space. During Stage~2, we jointly optimize the lightweight prompt factors and FiLM in \drift~, while keeping the new agent encoder and the ego fusion/head frozen. This design reduces the feature-domain gap without requiring access to raw sensor data or proprietary model metadata. Moreover, only the low-rank prompt factors and FiLM weights need to be stored for each modality, notably reducing storage compared to full perception stacks.

\subsection{Real-Time Deployment}
To support generalization, Stage~2 adaptation is performed offline for a set of common sensor/model configurations (referred to as \textit{modalities}). The ego agent stores the corresponding prompt--FiLM pair in \drift~for each modality. During real-time deployment, when a new agent joins, the ego must select the appropriate prompt--FiLM pair; an incorrect selection can lead to substantial performance degradation.

Many existing approaches perform this selection using sensitive sensor/model metadata, which may be unavailable due to privacy or proprietary constraints. Moreover, metadata alone can be insufficient. Even with the same sensor and backbone, a model trained on a different dataset exhibits a shifted feature distribution, causing a domain gap and degrading performance by selecting wrong \drift~modules. Hence, we developed a real-time, privacy-preserving mechanism to detect the joining agent's modality and select the appropriate modules without revealing sensitive metadata.

To identify the modality of a newly joining agent at runtime, \heteroprompt~extracts modality-discriminative cues from communicated BEV features using an AutoEncoder (AE). After completing Stage~1, we introduce an intermediate \textbf{Stage~1.5} to train the AE on ego domain features and construct a compact unified representation. In this stage, we keep the encoder, pyramid fusion, and detection head frozen, and train only the AE, as described in Algorithm~\ref{alg:deployment}. 

\begin{algorithm}[t]
\caption{Intermediate Stage~1.5 Training Logic}
\label{alg:deployment}
\begin{algorithmic}[1]

\Input{Ego feature $F_e$; aligned partner features
$\{F_{i\rightarrow e}\}_{i=1}^{N-1}$.}
\Output{Detection output $B$.}

\Statex \textbf{1.~Compact encoding and reconstruction:}

\State $z_e \gets g_{\text{AE}}(F_e)$
\Comment{Encode to compact latent}

\State $\bar{F}_e \gets d_{\text{AE}}(z_e)$
\Comment{Reconstruct unified feature}

\Statex \textbf{2.~Frozen fusion and detection:}

\State $H_{m_1}
\gets f_{\text{pyramid}}^{*}
\left(
F_{1\rightarrow e},
F_{2\rightarrow e},
\dots,
F_{N-1\rightarrow e},
\bar{F}_e
\right)$
\Comment{Fusion}

\State $B \gets f_{\text{head}}^{*}(H_{m_1})$
\Comment{Frozen head}

\State \Return $B$

\end{algorithmic}
\end{algorithm}

The AutoEncoder consists of an encoder $g_{AE}$ that maps the BEV features to a compact feature space, and a decoder $d_{AE}$ that maps the compact features back to the original size. $f^*$ is for frozen modules. 
Stage~1.5 is trained using the Stage~1 objective in~\eqref{eq:loss} together with a reconstruction loss:
\begin{equation}
\mathcal{L}_{1.5} = \mathcal{L}_{\text{Stage1}} + \lambda \left\|F_e - \bar{F}_e\right\|_2^2,
\end{equation}
where $\lambda$ balances detection and reconstruction. Therefore, the AE is guided by both the frozen Stage~1 detection pipeline and the reconstruction objective, encouraging the compact representation to preserve detection-relevant unified BEV features for real-time modality selection.

For a newly joining agent, its communicated BEV feature is first resized to the ego agent's unified resolution to ensure compatibility with the AE encoder. The AE encoder then maps the resized feature into a compact latent space, which captures modality-discriminative cues for identifying the agent's feature domain. We train a lightweight classifier on a collection of such compact latent codes obtained from a set of common modalities (sensor/model configurations).

During deployment, when a new agent joins, the ego agent applies the resizer, AE encoder, and classifier to infer the agent's modality as:
\begin{equation}
    \begin{gathered}
        z_m=g_{AE} (f_{resizer} (F_{m\rightarrow e}))\Rightarrow
        \Hat{m} = f_{cls}(z_m)
    \end{gathered}
\end{equation}
Based on the predicted modality ($\Hat{m}$), the ego selects the corresponding \drift~ modules, which enables real-time module selection without requiring any sensitive sensor or model metadata. After choosing the appropriate \drift s, the ego agent performs collaborative fusion and detection.

\section{EXPERIMENTS}
\subsection{Dataset and Heterogeneous Scenario Design}
We evaluated \heteroprompt~on two large-scale CP datasets: OPV2V-H~\cite{b12} for the main results and ablations, and V2XSet~\cite{b27} for further validation of the performance of our proposed framework. OPV2V-H~\cite{b12} is built upon OPV2V~\cite{b28} and is designed specifically for heterogeneous CP. Different perception encoders have been used both for LiDAR (\textsc{PointPillar}~\cite{b29} and \textsc{SECOND}~\cite{b30}) and camera (\textit{Lift-Splat-Shoot}~\cite{b31} to lift image features to BEV features, with \textsc{EfficientNet}~\cite{b32} and \textsc{ResNet-50}~\cite{b33} as image encoders) to evaluate the generalization capability of \heteroprompt~under heterogeneous CP settings.

For fair comparison with prior work, we define heterogeneous modalities following HEAL~\cite{b12}. Table~\ref{tab:modality} summarizes the four modalities used in our experiments. Modality $m_1$ is used as the ego agent to construct the ego-centric unified feature space with feature size $(64,128,256)$ using 64-channel LiDAR. We then progressively add three heterogeneous agents with modalities $m_2$, $m_3$, and $m_4$. All agents use pretrained, frozen single-agent encoders and backbones with $[0.4\,\mathrm{m}, 0.4\,\mathrm{m}]$ BEV grid. The detection range is $[-102.4\,\mathrm{m}, +102.4\,\mathrm{m}]$ and $[-51.2\,\mathrm{m}, +51.2\,\mathrm{m}]$ along $x$ and $y$.

\noindent\textbf{Implementation details.}
We use three ResNeXt~\cite{b25} blocks with $[3,5,8]$ layers in pyramid fusion to generate multi-scale features with $[64,128,256]$ channels, and $1\times1$ foreground estimators. The resizer applies average pooling to match the ego feature resolution. The FiLM module uses a three-layer CNN architecture, and the AutoEncoder contains three CNN layers, producing a compact latent feature of size $(64,8,16)$ from resized features. The classifier uses adaptive average pooling followed by three MLP layers. All training stages run for 35 epochs 
using the Adam optimizer. We report average precision (AP) at IoU thresholds of 0.5 and 0.7.
\begin{table}[t]
\centering
\caption{Heterogeneous modality definitions. The ego agent uses $m_1$ modality and $m_2$, $m_3$, and $m_4$ join progressively.}
\label{tab:modality}
\resizebox{0.9\columnwidth}{!}{
\begin{tabular}{l|c|c}
\toprule
\textbf{Modality} & \textbf{Sensor} & \textbf{Encoder} \\
\midrule
$m_1$ (ego) & LiDAR (64-channel) & \textsc{PointPillar} \\
$m_2$ & camera & \textsc{EfficientNet} \\
$m_3$ & LiDAR (32-channel) & \textsc{Second} \\
$m_4$ & camera & \textsc{ResNet-50}\\
\bottomrule
\end{tabular}}
\end{table} 

\subsection{Experimental Results Comparison}
We compare \heteroprompt~with state-of-the-art intermediate-fusion frameworks for heterogeneous collaborative perception under progressively added heterogeneous agents. As summarized in Table~\ref{tab:per}, \heteroprompt~consistently improves AP50 over HEAL~\cite{b12} on both datasets while using orders-of-magnitude fewer trainable parameters. On OPV2V-H, 
\heteroprompt~outperforms HEAL by $+1.8\%$, $+1.0\%$ and $+1.2\%$ at AP50, for $+m_2$, $+m_3$, and $+m_4$, respectively, while 
maintaining a comparable AP70. 
Additionally, \heteroprompt~on average improves  $+4.7\%/10.4\%$ and $+5.8\%/10.8\%$ at AP50/AP70, for CoBEVT~\cite{b34} and HM-ViT~\cite{b9}, respectively.
On V2XSet, \heteroprompt~similarly improves AP50 over HEAL by $+1.4\%$ for $+m_2$, $+1.5\%$ for $+m_3$ and $+1.1\%$ for $+m_4$, while maintaining a comparable AP70. Importantly, \heteroprompt~reduces the trainable parameters to only $0.064$M across all settings, yielding $94.1$--$99.6\%$ parameter reduction relative to HEAL, dramatically lower overhead than retraining baselines (CoBEVT~\cite{b34}, HM-ViT~\cite{b9}), supporting fast and privacy-preserving feature adaptation. 
We further evaluate training efficiency by comparing \heteroprompt~with HEAL. As reported in Table~\ref{tab:fast}, \heteroprompt~reduces peak GPU memory usage by $45.7\%$, while achieving end-to-end training throughput of $30.74$ samples/sec.\ ($2.5\times$ higher than HEAL) and $2.27\times$ higher computational throughput in TFLOPs/sec. on an NVIDIA RTX A6000 (batch size 1), demonstrating its computational efficiency.

\begin{table*}[!t]
\centering
\caption{Detection performance comparison on OPV2V-H (upper block) and V2XSet (lower block) using $m_1$ as the collaborative base and progressively adding three new heterogeneous agents ($m_2$, $m_3$, $m_4$). We report AP50\%, AP70\%, and the number of trainable parameters for No-Fusion (no retraining), Late Fusion, HEAL~\cite{b12}, CoBEVT~\cite{b34}, and HM-ViT~\cite{b9} ( retraining methods).}
\label{tab:per}

\resizebox{0.94
\textwidth}{!}{
\begin{tabular}{lccccccccc}
\toprule
\textbf{Based on $m_1$, Add New Agent} 
& \multicolumn{3}{c}{\textbf{$+m_2$}} 
& \multicolumn{3}{c}{\textbf{$+m_3$}} 
& \multicolumn{3}{c}{\textbf{$+m_4$}} \\
\cmidrule(lr){2-4} \cmidrule(lr){5-7} \cmidrule(lr){8-10}
\textbf{Metric \hspace{1cm} Dataset} 
& AP50 $\uparrow$ & AP70 $\uparrow$ & \# Params (M) $\downarrow$
& AP50 $\uparrow$ & AP70 $\uparrow$ & \# Params (M) $\downarrow$
& AP50 $\uparrow$ & AP70 $\uparrow$ & \# Params (M) $\downarrow$ \\
\midrule

No Fusion \hspace{0.5cm} OPV2V-H
& 74.8 & 60.6 & /   
& 74.8 & 60.6 & /   
& 74.8 & 60.6 & /   
\\

Late Fusion
& 77.5 & 59.9 & /   
& 83.3 & 68.5 & /   
& 83.4 & 68.5 & /   
\\

CoBEVT~\cite{b34}
& 82.2 & 67.1 & 33.2   
& 88.5 & 74.2 & 42.0   
& 88.5 & 74.2 & 51.7   
\\
HM-ViT~\cite{b9}& 81.3&64.6& 47.7 & 87.1& 74.3& 65.1& 87.6&75.5&83.3 \\

HEAL~\cite{b12}
& 87.2 & 78.3 & 15.0   
& 91.3 & \textbf{84.8} & 1.1    
& 90.9 & \textbf{84.5} & 1.9    
\\

\textbf{\heteroprompt~(ours)}
& \textbf{89.0} & \textbf{78.6} & \textbf{0.064 (99.6 $\%$ $\downarrow$)}   
& \textbf{92.3} & 84.2 & \textbf{0.064 (94.1 $\%$ $\downarrow$)}   
& \textbf{92.1} & 84.0 & \textbf{0.064(96.6 $\%$ $\downarrow$)}    
\\

\midrule
\midrule
HEAL~\cite{b12} \hspace{0.6cm} V2XSet
& 82.0 & 68.4 & 15.0   
& 82.8 & \textbf{71.2} & 1.1    
& 82.4 & \textbf{71.4} & 1.9   
\\
\textbf{\heteroprompt~(ours)}
& \textbf{83.4} & \textbf{68.7} & \textbf{0.064 (99.6 $\%$ $\downarrow$)}   
& \textbf{84.3} & 70.7 & \textbf{0.064 (94.1 $\%$ $\downarrow$)}   
& \textbf{83.5} & 70.5 & \textbf{0.064(96.6 $\%$ $\downarrow$)}    
\\
\bottomrule
\end{tabular}}
\end{table*}

\begin{table}[t]
\centering
\caption{Training computation cost comparison on OPV2V-H using heterogeneous scenario ($m_1+m_2+m_3+m_4$). We report training throughput (TT, samples/s), computational throughput (TF, TFLOPs/s), and peak GPU memory usage (PM).}
\label{tab:fast}
\resizebox{0.96\columnwidth}{!}{
\begin{tabular}{lccc}
\toprule
\textbf{Metric} & TT $\uparrow$ & TF $\uparrow$ & PM (GB) $\downarrow$ \\
\midrule
HEAL~\cite{b12} &12.34 &7.68&7.0\\
\textbf{\heteroprompt~(ours)} &  \textbf{30.74}\textbf{($\uparrow 2.5\times$)}&\textbf{17.44}\textbf{($\uparrow 2.27\times$)}&\textbf{3.8}\textbf{($\downarrow 45.7\%$)} \\
\bottomrule
\end{tabular}}
\end{table}
\subsection{Deployment Modality Classification}
\begin{figure}
    \centering
    \includegraphics[width=0.9\linewidth]{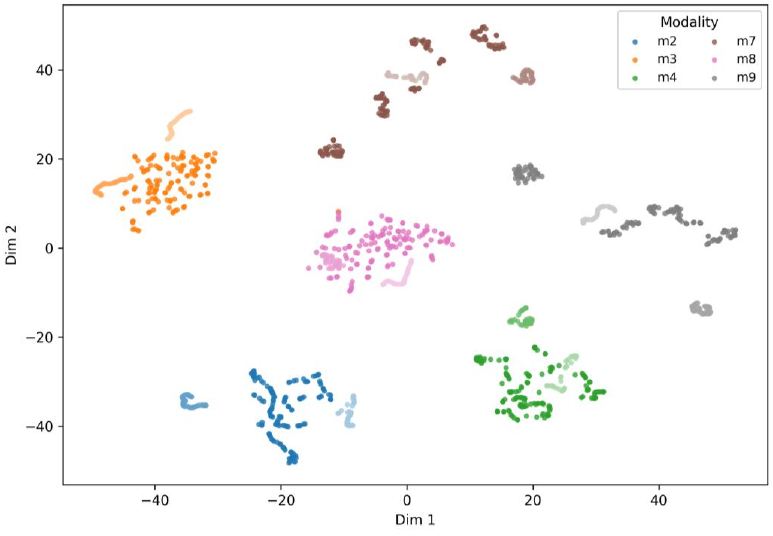}
     \caption{t-SNE visualization of AE compact features (flattened) for modality domains. $m_2$ to $m_4$ use OPV2V-H dataset and $m_7$ to $m_9$ use the same models with V2XSet dataset. Each color denotes a modality, and shade indicates data split (train/val/test). The compact space forms distinct modality clusters.}
     \label{fig:tsne}
\end{figure}

After Stage~1, we train an AutoEncoder (AE) to construct a compact unified feature space using the frozen ego encoder, fusion, and detection modules in an intermediate training stage. The AE encoder/decoder each use three CNN layers. We evaluate four latent sizes: $(64,8,16)$, $(64,16,32)$, $(128,8,16)$, and $(128,16,32)$. For $m_2$, $m_3$, and $m_4$, we extract compact features using 140/30/30 train/val/test samples per modality using OPV2V-H and V2XSet datasets. To visualize the geometry of the learned latent space, we apply t-SNE~\cite{b35} (perplexity 40--50) on flattened compact tensors. Fig.~\ref{fig:tsne} shows clear modality-dependent clustering in the compact space.

\begin{figure}
    \centering
    \includegraphics[width=0.85\linewidth]{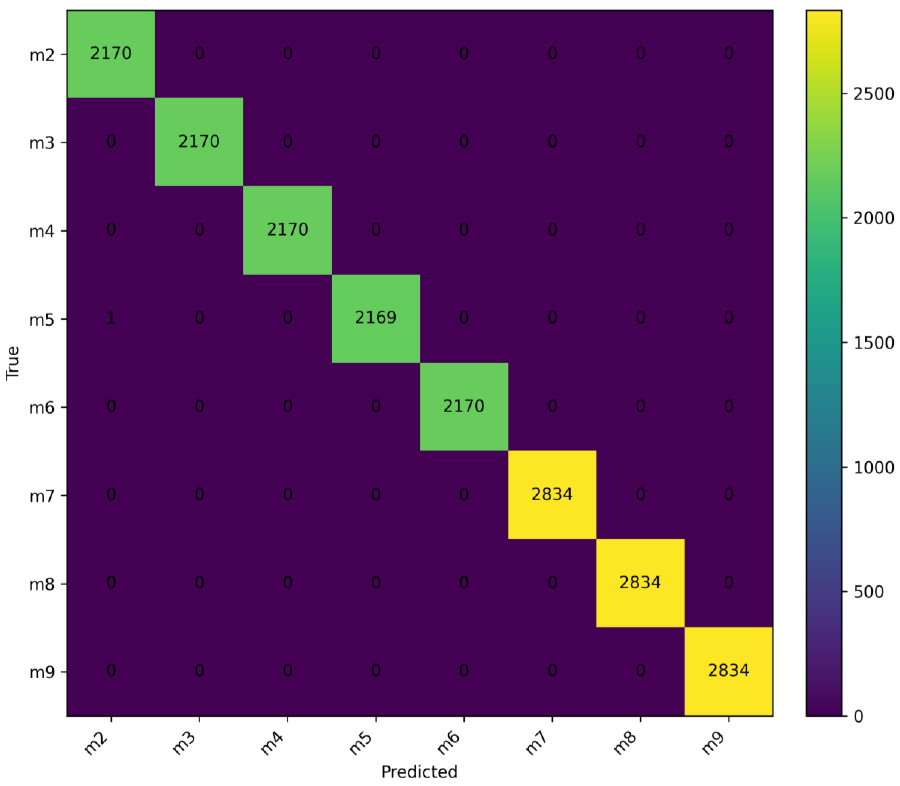}
    
\caption{Test confusion matrix for deployment modality classification using AE latent features and a lightweight MLP classifier. The classifier achieves $>99.99\%$ accuracy across $m_2$--$m_9$, with only one error between $m_5$ and $m_2$.}
    \label{fig:confusion}
\end{figure}

Table~\ref{tab:ae_latent} reports the classification test loss for different latent sizes. Since $(64,8,16)$ yields the lowest loss, we use it for classification with all samples. For modality classification, we apply average pooling on the compact features followed by a three-layer MLP classifier. To evaluate scalability, we further extend the modality set with $m_5$ (\textsc{SECOND}~\cite{b30}) and $m_6$ (\textsc{PointPillar}~\cite{b29}) with different voxel sizes than $m_3$ and $m_1$. Fig.~\ref{fig:confusion} reports the test confusion matrix over all modalities. The classifier achieves near-perfect accuracy ($>99.99\%$), with one confusion between $m_5$ and $m_2$.
\begin{table}[t]
\centering
\caption{AutoEncoder latent-size ablation for modality classification. Test loss is reported on a small sample for $m_2$ to $m_4$ (OPV2V-H), and $m_7$ to $m_9$ (same modality with V2XSet).}
\label{tab:ae_latent}
\resizebox{0.95\columnwidth}{!}{
\begin{tabular}{lcccc}
\toprule
\textbf{Metric} & $(64,8,16)$& $(64,16,32)$& $(128,8,16)$ &$(128,16,32)$ \\
\midrule
Test loss $\downarrow$     & \textbf{0.0290 }    & 0.0377     & 0.0858     & 0.0577      \\
\bottomrule
\end{tabular}}
\end{table}  

\subsection{Ablation Study}
We study the contribution of each \drift~module in Stage~2 heterogeneous adaptation of \heteroprompt. We used a CNN-based FiLM block with PARAFAC prompts of rank $R=16$, and evaluated three variants: using only \drift~1 (pre-resizer), only \drift~2 (post-resizer in the unified resolution), or both. Table~\ref{tab:ablation-study1} shows that combining both \drift~modules is important for effectively closing the feature-domain gap. Removing \drift~2 yields a drop of $2.9\%$ at AP50 and $3.4\%$ at AP70, while removing \drift~1 reduces performance by $4.6\%$/$4.8\%$ at AP50/AP70, indicating that \drift~1 contributes more strongly in this setting.
\begin{table}[t]
\centering
\caption{Effect of each \drift~module in Stage~2 feature-domain adaptation. \drift~1 has a larger contribution in this setting. Results are reported for $m_1+m_2+m_3+m_4$.}
\label{tab:ablation-study1}
\resizebox{0.77\columnwidth}{!}{
\begin{tabular}{ccccc}
\toprule
\textbf{\drift~1} &\textbf{\drift~2}& \textbf{AP50\% $\uparrow$}& \textbf{AP70\% $\uparrow$} \\
\midrule
\checkmark & \Crossmark & 89.2&80.6  \\
\Crossmark & \checkmark &  87.5& 79.2\\
\checkmark & \checkmark& \textbf{92.1}& \textbf{84.0} \\
\bottomrule
\end{tabular}}
\end{table}

\begin{figure}
    \centering
    \includegraphics[width=0.97\linewidth]{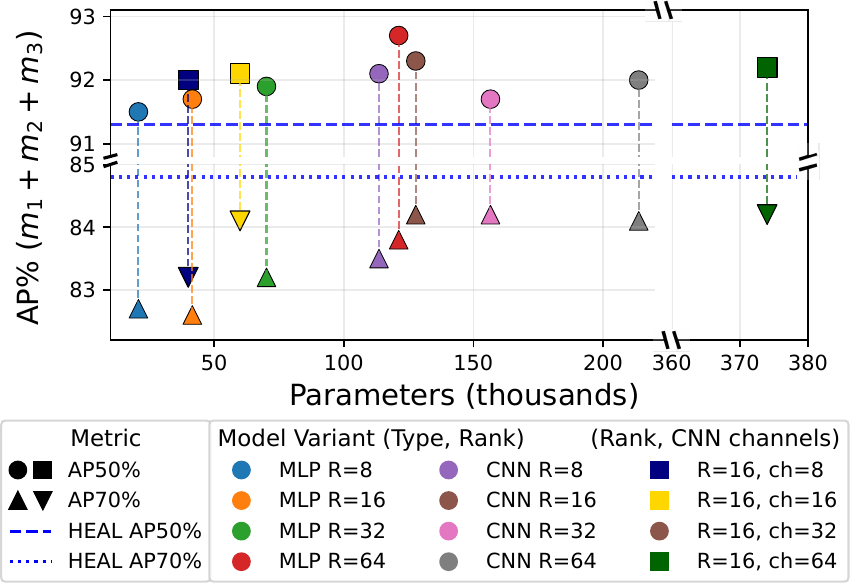}
    \caption{Accuracy-complexity ablations for Stage~2 new-agent adaptation under collaboration of three agents. We report AP50 and AP70 versus the number of trainable parameters. 1. Effect of PARAFAC prompt rank $R$ and FiLM architecture (MLP vs.\ CNN) ($\ocircle$:AP50 and $\triangle$:AP70). 2. For fixed $R=16$, effect of CNN FiLM hidden-layer channels ($\square$:AP50  and $\bigtriangledown$:AP70).}
    \label{fig:ablation2}
\end{figure}

We also evaluated the effect of the FiLM architecture (MLP vs.\ CNN) and the PARAFAC prompt rank $R$ on the accuracy--complexity trade-off under the $m_1+m_2+m_3$ setting. 
Fig.~\ref{fig:ablation2}
reports AP50 (circles) and AP70 (triangles) results versus the number of trainable parameters. Overall, CNN-based FiLM with $R=16$ (brown) provides the best AP70 performance while maintaining low complexity, and achieves competitive AP50 precision. Among the MLP architectures, a competitive option is MLP-based FiLM with $R=64$, which achieves the best AP50 performance. Increasing $R$ further increases model capacity and complexity with diminishing gains, while MLP-based FiLM generally yields lower complexity but weaker accuracy. As highlighted above, it is important to note that the HEAL AP results come at a high cost in complexity that is almost two orders of magnitude higher than \heteroprompt's overall complexity. For $R=16$, varying the CNN width (squares and upside-down triangles) shows that $32$ channels provides the best trade-off, whereas smaller widths reduce capacity and larger widths increase complexity with marginal gains.

\section{CONCLUSION}
In this paper, we introduced \heteroprompt, a heterogeneous collaborative perception framework for real-time, privacy-preserving, and rapid object alignment. Our method mitigates the BEV feature-domain gap across agents with different sensors and perception stacks by aligning communicated intermediate features to an ego-centric unified space using lightweight, prompt-conditioned FiLM interpreters. \heteroprompt~reduces trainable parameters by orders of magnitude while maintaining state-of-the-art detection performance by freezing the ego fusion/detection stack and joining-agent encoders, and adapting only two \drift~modules with PARAFAC low-rank prompts and feature-wise linear modulation. \heteroprompt~also preserves the privacy of joining agents by relying solely on received intermediate features. We further enable \textbf{metadata-free} and \textbf{privacy-preserving} deployment by learning a compact AutoEncoder representation and a classifier that infers the joining agent modality from features with $>99.99\%$ accuracy on the evaluated modality set, supporting real-time routing to the correct \drift~modules. Experiments on the OPV2V-H and V2XSet datasets demonstrate \heteroprompt’s efficiency in addressing the heterogeneous feature-domain gap without retraining whole perception stacks, providing a scalable and privacy-preserving approach for training and deployment. Future work will examine generalization to unseen sensors/models and robustness under stronger latency, packet loss, localization errors, and real-world V2X deployment.

\section{Acknowledgment}
This work was supported in part by the US Department of Transportation under grant 69A36523420190CRSCA.

\end{document}